# Uncertainty And Evolutionary Optimization: A Novel Approach

Maumita Bhattacharya
School of Computing & Maths
Charles Sturt University
Australia
maumita.bhattacharya@ieee.org

Rafiqul Islam
School of Computing & Maths
Charles Sturt University
Australia
mislam@csu.edu.au

Abdun Naser Mahmood
School of Engineering & IT
University of New South Wales
Canberra, Australia.
abdun.mahmood@unsw.edu.au

*Abstract*— Evolutionary algorithms (EA) have been widely accepted as efficient solvers for complex real world optimization problems, including engineering optimization [2, 3, 12, 13 and 17]. However, real world optimization problems often involve uncertain environment including noisy and/or dynamic environments, which pose major challenges to EA-based optimization. The presence of noise interferes with the evaluation and the selection process of EA, and thus adversely affects its performance. In addition, as presence of noise poses challenges to the evaluation of the fitness function, it may need to be estimated instead of being evaluated. Several existing approaches [4,5,6,7,8] attempt to address this problem, such as introduction of diversity (hyper mutation, random immigrants, special operators) or incorporation of memory of the past (diploidy, case based memory) [5 and 14]. However, these approaches fail to adequately address the problem. In this paper we propose a *Distributed Population Switching Evolutionary Algorithm (DPSEA)* method that addresses optimization of functions with noisy fitness using a distributed population switching architecture, to simulate a distributed self-adaptive memory of the solution space. Local regression is used in the pseudo-populations to estimate the fitness. Successful applications to benchmark test problems ascertain the proposed method's superior performance in terms of both robustness and accuracy.

## I. INTRODUCTION

Optimization problems involving noisy or uncertain fitness functions poses practical challenges as effectively the fitness function becomes a stochastic one in such cases. Many real world applications fall in this category, for example,

- **Online adaptation of real world systems**: In such systems some design parameters can be decided only through experiments and in real time.
- **Simulation based optimization of large and complex systems**: Simulations that use random numbers to represent a stochastic process in the simulation. Use of random numbers causes changes in the fitness value obtained by simulation.

A noisy environment in the current context may be viewed as one where the fitness of the solution is disturbed, i.e. if $f(x)$ is the fitness function and $\delta$ is some (e.g. Gaussian) noise, then a noisy fitness function would mean $F(x) = f(x) + \delta$ and the optimum of $f(x)$ is to be evaluated despite the noise. This noise is assumed to be independent of the location of the individuals.

Presence of noise in the fitness function could result in 'overestimation' of fitness of inferior candidates and similarly 'underestimation' of fitness of superior candidates during the selection process. This is likely to result in reduced learning rate, inability to retain learnt information, limited exploitation and absence of gradual monotonous improvement of fitness with generations.

The effects of noise can be reduced using re-sampling and large population size. However, these simple methods are impractical within a limited time frame (interpreted as limits on the total number of fitness evaluations) as fitness evaluation can be highly expensive in many real life problem domains. Also excessive re-sampling is rather likely to deteriorate performance if the population size is small.

Some of the key techniques that have been used to tackle the problem of uncertainty in EA are: *Re-sampling* [6, 4, 5], *Conventional EA with increased population size* [4, 6], *Rescaled mutation* [1], *Thresholding* [6, 5], *Fitness value based on neighbouring individuals* [6, 9, 3, and 8] and *Reduced resampling* or *partial resampling* [7, 6, 12 and 13]. A comprehensive survey of various techniques to handle noisy environment with EA, including multipopulation approaches can be found in [5 and 14].

In this paper we focus on *uncertain* problem domains where accuracy and robustness of solution are more important compared to sample size for resampling or cost of fitness evaluation. The proposed DPSEA framework involves distributed population switcing architecture with local regression employed in the sub-populations. DPSEA framework aims at generating relatively low cost, accurate and robust solution in uncertain environment involving noise.

The rest of the paper is organized as follows: Section II details the proposed DPSEA framework. Details of experiments are given in Section III, while Section IV presents the results and discussions. Finally, Section V concludes the paper.

## II. THE PROPOSED ALGORITHM STRUCTURE

### A. The Operational Principle of DPSEA

The proposed DPSEA framework uses a hybrid single to multi-population switching architecture with partial local regression in the pseudo-populations. Distribution of the population allows tracking multiple peaks in the search space. Each pseudo-population maintains information about a separate region in the search space, thus acting as a distributed self adaptive memory. We use the term pseudo-population instead of subpopulation to emphasize on the fact that each one of these sub groups may not actually act as self-sufficient evolving populations. The functioning of the proposed model is guided by the following key notions:

- Retention of memory with distributed pseudo-populations in the search space, i.e., the pseudo-populations should be able to track their moving peaks over a period.
- Maintenance of diversity through pseudo-populations.
- Estimation of noisy fitness by partial local regression in each pseudo-population. Quadratic regression is used in the present work. However, linear regression could be used to reduce computation time.

The basic algorithm structure of DPSEA is as described in Fig. 1.

### B. The Mechanism Involved

Specific features of the proposed distributed population switching EA are as follows:

- Unlike many conventional multipopulation EAs [5] the distributed population switching EA does not maintain a main population alongside the pseudo-populations. Instead, DPSEA switches from single population to multi-population periodically.
- At *switching generation* step, the pseudo-populations are merged in to regain the main population and evolution is carried out as per canonical GA mechanism. To correct the adverse effects of noise, actual fitness evaluation along with resampling is used at this stage. Considering the infrequency of this step, the cost is not overwhelming.
- The regenerated main population then dissolves into pseudo-populations by self-organization. This is essentially distribution of the candidate solutions into pseudo-populations based on specific criteria. For example, fitness and *population size* factors have been used to decide the eligibility of a pseudo-population to obtain evolution right. An eligible pseudo-population then evolves by the canonical GA mechanism. Partial local regression is used to estimate the fitness values. Mutation rate depends on a factor of *fitness* and *population size* as well.
- Members of the non-eligible pseudo-populations either survive or eventually disappear following the *switching generations*.

- Retention of memory about the *moving peaks* is achieved through the pseudo-populations over specific durations until the *switching generation* step. This is logical considering the uncertain nature of the phenotypic solution space.

As in [4] estimation of the fitness by local regression is based on the following assumptions:

- $f(x)$ can be locally approximated by a low polynomial function (linear or quadratic).
- The variance within a local region (pseudo-population in this case) is constant.
- The noise is normally distributed, i.e. $\delta(x) \sim N(0, \sigma^2(x))$.

The simulation details for DPSEA are presented in section III and section IV.

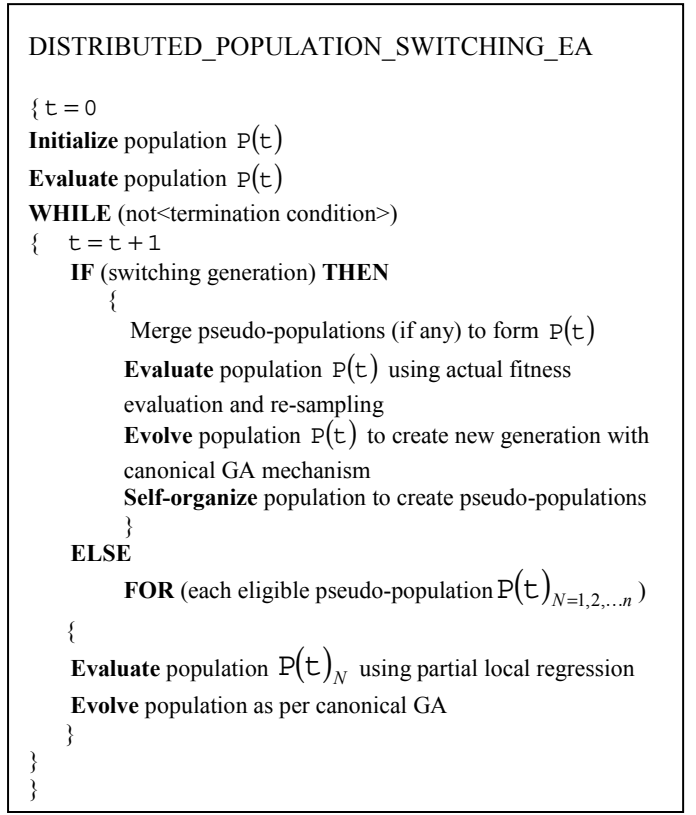

```
DISTRIBUTED_POPULATION_SWITCHING_EA

{ t = 0
Initialize population P(t)
Evaluate population P(t)
WHILE (not<termination condition>)
{   t = t + 1
    IF (switching generation) THEN
       {
         Merge pseudo-populations (if any) to form P(t)
         Evaluate population P(t) using actual fitness
         evaluation and re-sampling
         Evolve population P(t) to create new generation with
         canonical GA mechanism
         Self-organize population to create pseudo-populations
       }
    ELSE
       FOR (each eligible pseudo-population P(t)_{N=1,2,...n} )
    {
    Evaluate population P(t)_N using partial local regression
    Evolve population as per canonical GA
    }
}
}
```

Fig. 1. The Distributed Population Switching EA architecture.

## III. SIMULATION DETAILS

DPSEA aims at improving the accuracy and robustness of solution in a noisy environment. To this end, in our experiments, we investigated the performance of the proposed DPSEA model when applied to a set of standard benchmark functions. In this research, we also investigated the performance of the proposed DPSEA model as against standard GA, differential evolution, particle swarm optimization and a variant of ant colony optimization for noisy benchmark functions (see Table II). For the purpose of analysis, experiments have been conducted on the non-noisy versions of the same set of benchmark functions as well.

## A. Test Functions

We have used the following benchmark functions to test the algorithms: Sphere function (5D), Griewank's function (50D), Rastrigin's function 1 (50D) and the Rosenbrock's function (50d). These benchmark functions have previously been used by researchers to test performance of evolutionary algorithm for noisy optimization problems [3]. All the benchmarks used are minimization problems. The descriptions of the functions are as below:

Sphere function (5 Dimensional):

$$f_{Sphere}(\vec{x}) = \sum_{i=1}^{5} x_i^2 \text{ with } -100 \leq x_i \leq 100 \quad (1)$$

Griewank's function (50 Dimensional):

$$f_{Griewank}(\vec{x}) = \frac{1}{4000} \sum_{i=1}^{50} (x_i - 100)^2$$
$$- \prod_{i=1}^{50} \cos\left(\frac{x_i - 100}{\sqrt{i}}\right) + 1 \text{ with } -600 \leq x_i \leq 600 \quad (2)$$

Rastrigin's Function 1 (50 Dimensional):

$$f_{RastriginF1}(\vec{x}) = 200 + \sum_{i=1}^{50} x_i^2 - 10 \cdot \cos(2\pi x_i)$$
with $-5.12 \leq x_i \leq 5.12$ (3)

Rosenbrock's function (50 Dimensional):

$$f_{Rosenbrock}(\vec{x}) = \sum_{i=1}^{49} \left[100(x_{i+1} - x_i^2)^2 + (x_i - 1)^2\right]$$
with $-50 \leq x_i \leq 50$ (4)

The noisy versions of the above set of functions are defined as:

$$f_{Noisy}(\vec{x}) = f(\vec{x}) + N(\mu, \sigma^2) \quad (5)$$

where, $N(\mu, \sigma^2)$ = Standard Normal (or Gaussian) distribution with mean, $\mu = 0$ and variance, $\sigma^2 = 1$. The probability density function $f(x; \mu, \sigma^2)$ is given as below,

$$f(x; \mu, \sigma^2) = \frac{1}{\sigma\sqrt{2\pi}} \exp\left(-\frac{(x-\mu)^2}{2\sigma^2}\right) \quad (6)$$

## B. Experiment Settings

The parameter settings for the algorithms are as shown in Table I. For comparison purpose we have used the methods reported in [7] for conventional GA, differential evolution (DE) and particle swarm optimization (PSO) and the recent Multilevel Ant Stigmergy Algorithm (MASA) reported in [16]. Similar parameter settings have been used to ensure equality of comparison. As the same parameter setting has been used for all the four functions, the parameters of the heuristic algorithms were not tuned for individual test problem. This is reasonable considering the fact that in case of real life problems such tuning can be infeasible due to time constraints.

Let $totalEval$ = total number of function evaluations, $popSize$ = population size, $totalIT$ = total number of iterations, $rs$ = total number of candidate solution resampling, and $totalIUnchanged$ =total number of individuals that remained unchanged (such as the elites) and were not reevaluated in the various generations. Then,

$$totalEval = popSize * totalIT * rs - totalIUnchanged \quad (7)$$

The total number of function evaluations in [7] in case of conventional GA, differential evolution and PSO is kept constant by keeping $totalIT \propto \frac{1}{rs}$. However, this has not been followed for the proposed method. We have used a fixed number of iteration in these experiments. The experiments were run with different values of $rs$ such as $rs$ =1, 5, 20, 50 and 100, to eradicate the effects of noise and find the 'true' fitness. The 'true' fitness here refers to the fitness value obtained by evaluation of the non-noisy versions of the same functions.

TABLE I : PARAMETER SETTINGS* OF THE ALGORITHMS

| Proposed Method | Canonical GA | Differential Evolution | PSO |
|---|---|---|---|
| popSize=100 | popSize=100 | popSize=50 | popSize=20 |
| $p_c$=1.0 | $p_c$=1.0 | $CF$=0.8 | $w$= 1.0→0.7 |
| $p_m$=0.3 | $p_m$=0.3 | $f$=0.5 | $\varphi_{min}$=0.0 |
| $n$=10 | $n$=10 | | $\varphi_{max}$=2.0 |
| $\sigma_m$=0.01 | $\sigma_m$=0.01 | | |

*popSize=Population Size, $p_c$= Crossover rate (EA), $p_m$= Mutation Rate (EA), $n$= Number of Elites, $\sigma_m$= Mutation Variance, $CF$= Crossover Factor (in differential Evolution), $f$=Scaling Factor, $w$=Inertia Weight (the value is linearly decreased from 1.0 to 0.7 during the run), $\varphi_{min}$, $\varphi_{max}$ = Lower and Upper Bounds of the Random Velocity Rule Weights.

## IV. REASULTS AND DISCUSSIONS

### A. Performance of DPSEA

Table II present the final results obtained for the chosen benchmark functions (both noisy and non-noisy versions). The results for the non-noisy versions of the functions have been reported here mainly as 'baseline' to judge the impact of noise. For comparison purpose we have used the results reported in [7] as well as the recent outcome [16] for the following methods: differential evolution, particle swarm optimization (PSO) and conventional EA (CGA). As mentioned earlier, the total number of function evaluations has been kept fixed for the experiments with differential evolution, PSO, conventional EA [7] and MASA [16]. The proposed method used variable number of function evaluations.

The total number of function evaluations used for the various methods have been as reported in Table III.

TABLE II: PERFORMANCE (AVERAGE BEST FITNESS) COMPARISON FOR DPSEA, CGA, DIFFERENTIAL EVOLUTION, PSO AND MASA. $rs$ =NUMBER OF CANDIDATE SOLUTION RESAMPLING.

| | | DPSEA | CGA | Differential Evolution | PSO | MASA |
|---|---|---|---|---|---|---|
| $f_1$ =Sphere function (5D non-noisy) And $f^*_1$ =Sphere function (5D noisy) | $f_1$ | 4.21334E-75±0 | 6.71654E-20±0 | 4.12744E-152±0 | 2.51130E-8±0 | 0 (50 D) |
| | $f^*_1, rs=1$ | 0.00103E-2±0.003 | 0.04078±0.00543 | 0.25249±0.02603 | 0.36484±0.05182 | - |
| | $f^*_1, rs=5$ | 0.00023E-2±0.001 | 0.02690±0.00363 | 0.13315±0.01266 | 0.16702±0.03072 | -0.328 (50 D) |
| | $f^*_1, rs=20$ | 0.00018E-2±0.023 | 0.02205±0.00290 | 0.07364±0.00811 | 0.11501±0.01649 | 10.545 (50 D) |
| | $f^*_1, rs=50$ | 0.00001E-2±0.0031 | 0.01765±0.00233 | 0.07004±0.00686 | 0.06478±0.00739 | 76.441 (50 D) |
| | $f^*_1, rs=100$ | 0.00011E-2±0.0009 | 0.03929±0.00396 | 0.08165±0.00800 | 0.07135±0.00938 | 1062.354 (50 D) |
| $f_2$ =Griewank function (50D non-noisy) And $f^*_2$ =Griewank function (50D noisy) | $f_2$ | 4.00E-7±0.001 | 0.00624±0.00138 | 0±0 | 1.54900±0.06695 | 0 |
| | $f^*_2, rs=1$ | 0.00211E-1±0.001 | 1.14598±0.00307 | 3.31514±0.07388 | 11.2462±0.50951 | - |
| | $f^*_2, rs=5$ | 0.00211E-1±0.001 | 1.10223±0.00342 | 2.42183±0.03616 | 16.6429±0.70800 | 0.132 |
| | $f^*_2, rs=20$ | 0.00011E-1±0.021 | 1.44349±0.01381 | 2.67093±0.03895 | 85.4865±2.13148 | 1.898 |
| | $f^*_2, rs=50$ | 0.00200E-1±0.323 | 3.69626±0.13127 | 46.8197±0.96449 | 143.021±2.33228 | 2.946 |
| | $f^*_2, rs=100$ | 0.00301E-1±0.481 | 18.0858±0.99646 | 233.802±6.25840 | 194.188±4.90959 | 17.040 |
| $f_3$ =Rastrigin's function 1 (50D non-noisy) And $f^*_3$ =Rastrigin's function 1 (50D noisy) | $f_3$ | 1.219E-7±0.0013 | 32.6679±1.94017 | 0±0 | 13.1162±1.44815 | 0 |
| | $f^*_3, rs=1$ | 0.03011E-1±0.031 | 30.7511±1.32780 | 2.35249±0.06062 | 55.9704±2.19902 | - |
| | $f^*_3, rs=5$ | 0.01071E-1±0.011 | 31.4725±2.02356 | 14.0355±0.47935 | 160.500±2.67500 | 2.132 |
| | $f^*_3, rs=20$ | 0.01171±0.111 | 39.1777±2.11529 | 167.628±2.12569 | 313.184±3.93659 | 10.238 |
| | $f^*_3, rs=50$ | 0.01092±0.100 | 74.8577±2.69437 | 314.762±2.88650 | 380.178±4.88706 | 22.213 |
| | $f^*_3, rs=100$ | 0.01232±0.190 | 147.800±2.93208 | 438.036±3.67504 | 418.265±5.35434 | 114.638 |
| $f_4$ =Rosenbrock's function (50D non-noisy) And $f^*_4$ =Rosenbrock's function (50D noisy) | $f_4$ | 1.4011E-14±0.0011 | 79.8180±10.4477 | 35.3176±0.27444 | 5142.45±2929.47 | 0.744E-1 |
| | $f^*_4, rs=1$ | 1.4201E-4±0.0021 | 118.940±13.2322 | 47.6188±0.15811 | 4884.68±886.599 | - |
| | $f^*_4, rs=5$ | 1.1201E-4±0.0091 | 341.788±49.6738 | 47.0404±0.13932 | 368512±39755.5 | 25.348 |
| | $f^*_4, rs=20$ | 1.0001±0.2191 | 1859.06±261.844 | 7917.46±352.851 | 1.61E+7±1.18E+6 | 214.532 |
| | $f^*_4, rs=50$ | 1.0091±0.1393 | 35477.7±4656.17 | 1.65E+7±903677 | 5.57E+7±2.38E+6 | 3051.539 |
| | $f^*_4, rs=100$ | 2.0028±2.1999 | 257488±19371.2 | 2.98E+8±1.04E+7 | 1.17E+8±7.38E+6 | 1270827 |

A comparison of total number of function evaluations is given in Table III.

It can be seen, all the heuristics have performed considerably better on the non-noisy benchmark functions compared to the noisy versions of the same functions. While the proposed method has performed better for majority of the test cases (see Table III), in terms of number of functions evaluaiton, the difference is not necessarily significant in all cases of the low dimensional Sphere function. However, for the high dimensional (50D) Griewank, Rastrigin and Rosenbrock functions the differences are clearly significant. In the remaining part of this section we will focus our discussions to performances related to the noisy functions.

Similar to the observation reported in [7], resampling has prevented stagnation in all the four test functions (see Table II), regardless of whether the mean final result has been improved. However, performance deteriorated with increased degree of sampling in case of MASA [16]. However, the effect of rate of resampling is inconclusive and may be problem dependent,, since it varies from case to case . Furthermore,  it is obvious that in case of conventional EA, increased number of resampling cannot improve the performance when number of iterations is inversely proportional to number of resampling to keep the total number of function evaluations constant. Interestingly for most of the test cases moderate rate of resampling has helped to improve the solution (see Table II), while high to very high rate of resampling has rather a deteriorating effect on the solutions.

TABLE III: TOTAL NUMBER OF FUNCTION EVALUATIONS USED TO ATTAIN THE REPORTED RESULTS WITH DPSEA, CANONICAL GA, DIFFERENTIAL EVOLUTION, PSO AND MASA FOR THE VARIOUS TEST CASES.

| Method | $totalEval$ for 5D Sphere function | $totalEval$ for 50D Griewank function | $totalEval$ for 50D Rosenbrock and Rastrigin function |
|---|---|---|---|
| DPSEA | 90000 | 430,000 | 450,000 |
| CGA | 100,000 | 500,000 | 500,000 |
| Differential Evolution | 100,000 | 500,000 | 500,000 |
| PSO | 100,000 | 500,000 | 500,000 |
| MASA | 500,000 (50 D) | 500,000 | 500,000 |

As can be observed from Table III, the proposed model has managed to produce far superior solutions with much fewer actual function evaluations.

The success rates of the DPSEA algorithm for simulations with noisy functions for various values of noise's standard deviation $\sigma$ are summarized in Table IV. Simulations were repeated 100 times for each noise standard deviation $\sigma$ value. The reported results are for the $rs=1$ simulation cases. Here 'success' has been defined slightly differently for the different test functions.

TABLE IV: ANALYSIS OF SUCCESS RATES OF THE DPGA ALGORITHM FOR THE NOISY VERSIONS OF THE TEST FUNCTIONS WITH VARIOUS VALUES OF NOISE'S STANDARD DEVIATION $\sigma$.

| $\sigma$ | $f_1^*$ $rs=1$ | $f_2^*$ $rs=1$ | $f_3^*$ $rs=1$ | $f_4^*$ $rs=1$ |
|---|---|---|---|---|
| 0 | 100% | 100% | 100% | 100% |
| 0.1 | 75% | 70% | 65% | 100% |
| 0.2 | 88% | 85% | 75% | 100% |
| 0.3 | 88% | 75% | 75% | 95% |
| 0.4 | 76% | 70% | 65% | 95% |
| 0.5 | 58% | 55% | 58% | 95% |
| 0.7 | 40% | 50% | 53% | 90% |
| 0.9 | 75% | 70% | 69% | 95% |

*B. Comparison of Estimation Approaches*

The estimation approach used in DPSEA has some similarity to the ones used in [ 4,9 and 11]. In Table V we have presented the comparative settings for various defining parameters of the estimation approaches used in these four techniques.

TABLE V: COMPARISON OF DIFFERENT ESTIMATION APPROACHES.

| | Proposed Model | [4] | [9] | [11] |
|---|---|---|---|---|
| Estimation | linear, quadratic | constant, linear, quadratic | constant | constant |
| Model | local regression in pseudo-populations | local regression | increasing variance | local regression |
| Locality | neighborhood | neighborhood self-adaptive, optimized | global | fixed neighborhood |
| Weighting | | sigmoidal weight function | equal for all samples | linear weight function |
| Variance | locally constant | locally constant | linear, optimized | locally constant |

## V. CONCLUSIONS

Many real world problems often involve complex optimization. Optimization problems involving uncertain environments are challenging to evolutionary algorithms as they require finding the optimum, where the problem operates in noisy environment (noisy fitness function in the present case). Numerous methods, including the multipopulation approach, special operators, case-based memory and so on have been tried to face this challenge with varied degrees of success. The proposed Distributed Population Switching EA (DPSEA) framework presented in this paper is comparable to the multipopulation approach in that it divides the search or solution space into multiple pseudo-populations and retains fitness information in them over a period. However, the DPSEA algorithm applies superior mechanism to reduce the computational expense of maintaining the main population along with the subpopulations by switching between main and subpopulations at regular intervals. Furthermore, DPSEA approach allows enough retention of memory so that dissolving the pseudo-populations periodically do not have any adverse effect. Compared to many existing approaches (such as [4] and [16]), DPSEA is conceptually straightforward. It does not involve determination of ambiguous weight function. The simulation results for DPSEA have been promising ascertaining the effectiveness of the proposed algorithm in the chosen problem domain.